\documentclass[letterpaper, 10 pt, conference]{ieeeconf}  

\IEEEoverridecommandlockouts                              

\usepackage{graphics} 
\usepackage{epsfig} 
\usepackage{amsmath} 
\usepackage{amssymb}  
\usepackage[table,xcdraw,dvipsnames]{xcolor}
\usepackage{multirow}
 \usepackage{booktabs}
\usepackage{multirow}
\usepackage{algorithm}
\usepackage{algorithmic}
\usepackage{hyperref}       

\usepackage[shrink=10,letterspace=500]{microtype}

\definecolor{agg_g}{RGB}{112,173,71}
\definecolor{agg_r}{RGB}{255,0,0}

\usepackage{subcaption}

\graphicspath{{images/}}

\title{Getting SMARTER for Motion Planning in Autonomous Driving Systems}
\author{Montgomery Alban$^1$, Ehsan Ahmadi$^{1,2}$, Randy Goebel$^2$, Amir Rasouli$^1$
\thanks{$^{1}$Huawei Technologies Canada}
\thanks{{\tt montgomery.alban@h-partners.com}}%
\thanks{{\tt amir.rasouli@huawei.com}}%
\thanks{$^{2}$University of Alberta, {\tt\ \{eahmadi,rgoebel\}@ualberta.ca}}}

\begin{document}
\maketitle
\begin{abstract}
Motion planning is a fundamental problem in autonomous driving and perhaps the most challenging to comprehensively evaluate because of the associated risks and expenses of real-world deployment. Therefore, simulations play an important role in efficient development of planning algorithms. To be effective, simulations must be accurate and realistic, both in terms of dynamics and behavior modeling, and also highly customizable in order to accommodate a broad spectrum of research frameworks. In this paper, we introduce SMARTS 2.0, the second generation of our motion planning simulator which, in addition to being highly optimized for large-scale simulation, provides many new features, such as realistic map integration, vehicle-to-vehicle (V2V) communication, traffic and pedestrian simulation, and a broad variety of sensor models.

Moreover, we present a novel benchmark suite for evaluating planning algorithms in various highly challenging scenarios, including interactive driving, such as turning at intersections, and adaptive driving, in which the task is to closely follow a lead vehicle without any explicit knowledge of its intention. Each scenario is characterized by a variety of traffic patterns and road structures. We further propose a series of common and task-specific metrics to effectively evaluate the performance of the planning algorithms. At the end, we evaluate common motion planning algorithms using the proposed benchmark and highlight the challenges the proposed scenarios impose. The new SMARTS 2.0 features and the benchmark are publicly available at \href{https://github.com/huawei-noah/SMARTS}{github.com/huawei-noah/SMARTS}.
\end{abstract}

\section{Introduction}
One of the key challenges in autonomous driving (AD) is motion planning, i.e., generating behavior for an autonomous vehicle to navigate safely in a highly interactive and stochastic environment. Unlike other aspects of AD systems, such as perception and prediction, evaluation of motion planning algorithms in the real world may not be feasible due to potential safety concerns, high cost of deployment, and the inability to reproduce scenarios for repeated testing and comparison of solutions. As a result, simulators have become integral tools in the development of motion planning algorithms. To be effective, simulators must be comprehensive, addressing every aspect of real-world driving, and should be realistic to accurately model dynamics, behaviors, observations that correspond to real-world environments  \cite{zhang2022rethinking}. 

In this paper, we introduce SMARTS 2.0, which is built upon our successful and evolving simulation platform \cite{zhou2020smarts}, with the intention of bringing together realistic behavioral modeling, efficiency, and diagnostic capabilities for effective development of motion planning algorithms. In addition to the original features of SMARTS, the new version provides support for integration of real-world datasets, sensor simulation, baseline evaluation protocol, vehicle-to-vehicle (V2V) communication, and diagnostic tool-sets, all of which are implemented in a highly optimized fashion, allowing large-scale simulation of heterogeneous agents.

Furthermore, we propose a novel motion planning benchmark, which focuses on highly interactive scenarios, such as turning actions at complex intersections and vehicle following maneuvers. As part of our contribution, we propose novel metrics and evaluate a number of existing planning algorithms, and highlight their challenges to motivate future research.

\section{Related Work}
\subsection{AD Simulation}
Given the associated costs and lack of reproducibility in real-world scenarios, simulation serves as a fundamental tool for developing and evaluating motion planning algorithms. Increasing interest in autonomous driving has led to the introduction of many simulation platforms of varying scopes. Many of which cater to specific driving tasks, e.g. racing \cite{wymann2000torcs}, highway \cite{leurent2018environment}, or urban environment \cite{gulino2024waymax}. The focus of simulations also varies, whether it is on realistic sensor modeling \cite{muller2018sim4cv,dosovitskiy2017carla}, agent behavior modeling \cite{zhou2020smarts,sun2022intersim}, or execution efficiency \cite{gulino2024waymax,scibior2021imagining}.

The current landscape of simulation platforms offers various solutions for evaluating aspects of autonomous driving systems. Excluding some platforms, such as \cite{martinez2017beyond,leurent2018environment,amini2022vista}, which focus on planning in passive driving environments, the majority of simulations offer forms of behavior modeling for social agents (i.e., agents other than the ego-vehicle), which is often extended to multi-agent simulation. This allows multi-instance evaluation in single scenarios \cite{cai2020summit, palanisamy2020multi, craig_quiter_2020,santara2021madras, althoff2017commonroad}. Modeling sensors, such as visible spectrum cameras and LiDAR in some environments \cite{caesar2021nuplan,li2022metadrive, dosovitskiy2017carla, muller2018sim4cv, sun2022intersim, scibior2021imagining} enables the evaluation of full-stack autonomous driving systems. In some cases, occupancy maps are implemented via ray-casting, which provides a basis to model obstructions in the scene from the perspective of the ego-vehicle \cite{vinitsky2022nocturne}. Some platforms also offer unique features, such as mechanisms for V2V communication \cite{palanisamy2020multi}. More recent simulators emphasize behavioral realism \cite{xu2023bits, gulino2024waymax, vinitsky2022nocturne, li2022metadrive} by incorporating expert data (in the form of human demonstrations or trajectories from expert policies) and real-world traffic data, which helps to reduce the sim-to-real gap in training driving models. For a more detailed list of simulations see \cite{Li_2024_tiv}.

In this work, we introduce SMARTS 2.0, built upon the widely used SMARTS \cite{zhou2020smarts} simulator, with a goal of integrating many of the aforementioned features into a single platform. This provides an improved foundation for development of a comprehensive study of AD motion planning algorithms. Our new platform provides a tool-set for the realistic simulation of heterogeneous traffic agents, sensors, and communication channels. In addition, it provides diagnostic tools for evaluating algorithms and further optimizes agent interaction for large-scale multi-agent simulation.

\subsection{Motion Planning Benchmarks}
There are a number of existing benchmarks, e.g. \cite{houston2021one, caesar2020nuscenes, Wilson_Argoverse2}, that focus on the prediction task, which is a key component for motion planning. Here, the goal is to accurately measure the behavior of the social agents, often using a variety of accuracy or diversity metrics \cite{chen2024criteria}. 

Motion planning benchmarks \cite{rasouli2023driving, althoff2017commonroad, Sun_2020_CVPR, Ansys_comp, TPCAP_comp} take one further step and model the response of the mission vehicle to the evolving surrounding environment. Specialized benchmarks, however, evaluate other aspects of the generated behaviors. For example, the CARLA benchmark \cite{dosovitskiy2017carla} is based on synthetic scenarios inspired by the US National Highway Traffic Safety Administration (NHTSA) pre-crash typology whose evaluation is based on route and infraction points. The CommonRoad Benchmark \cite{althoff2017commonroad} focuses on both interactive and non-interactive environments (where trajectories of social agents are provided) and assesses driving styles and rule violations with no restrictions. The Waymo simulation benchmark \cite{Sun_2020_CVPR} is intended for generating human-like trajectories, which are evaluated against real data in terms of realism and distributional distance error. 

In this work, we propose a novel benchmark, focusing on highly interactive scenarios which involve both turning actions at complex intersections and adaptive driving in car following tasks. For evaluation, we improve upon the existing metrics and propose new ones to better capture the practicality of algorithms for real-world applications.

\section{SMARTS 2.0}
SMARTS \cite{zhou2020smarts} is an open-source platform designed for research on motion planning for autonomous driving. Some of the key features of SMARTS include a compositional architecture and distributed computing for scalable simulation, highly realistic physics engine, and support for standard Gym\footnote{https://gymnasium.farama.org/} APIs for improving the flexibility of creating custom environments. Despite the popularity of the platform, our first version had several shortcomings, including the lack of an established evaluation protocol, limited support for data ingratiation, and a number of limitations on context modeling and sensing. In SMARTS 2.0 we address many of these prior shortcomings. Below are the descriptions of new key features and improvements in our simulation platform.
\begin{figure}
    \centering
    \includegraphics[width=1\linewidth]{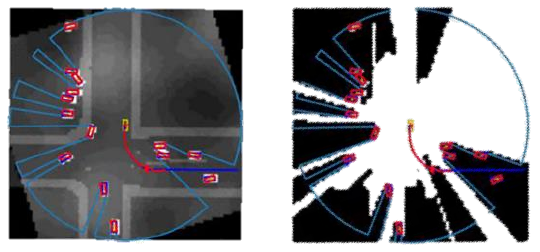}
    \caption{Left to right: bird-eye-view image of the simulation environment, and the visualization of the 2D sensor observation space. The mission vehicle is centered in the map and the blue circle shows the maximum observation range of the sensor.}
    \label{fig:sensor}
\end{figure}

\begin{figure}
    \centering
    \includegraphics[width=1\linewidth]{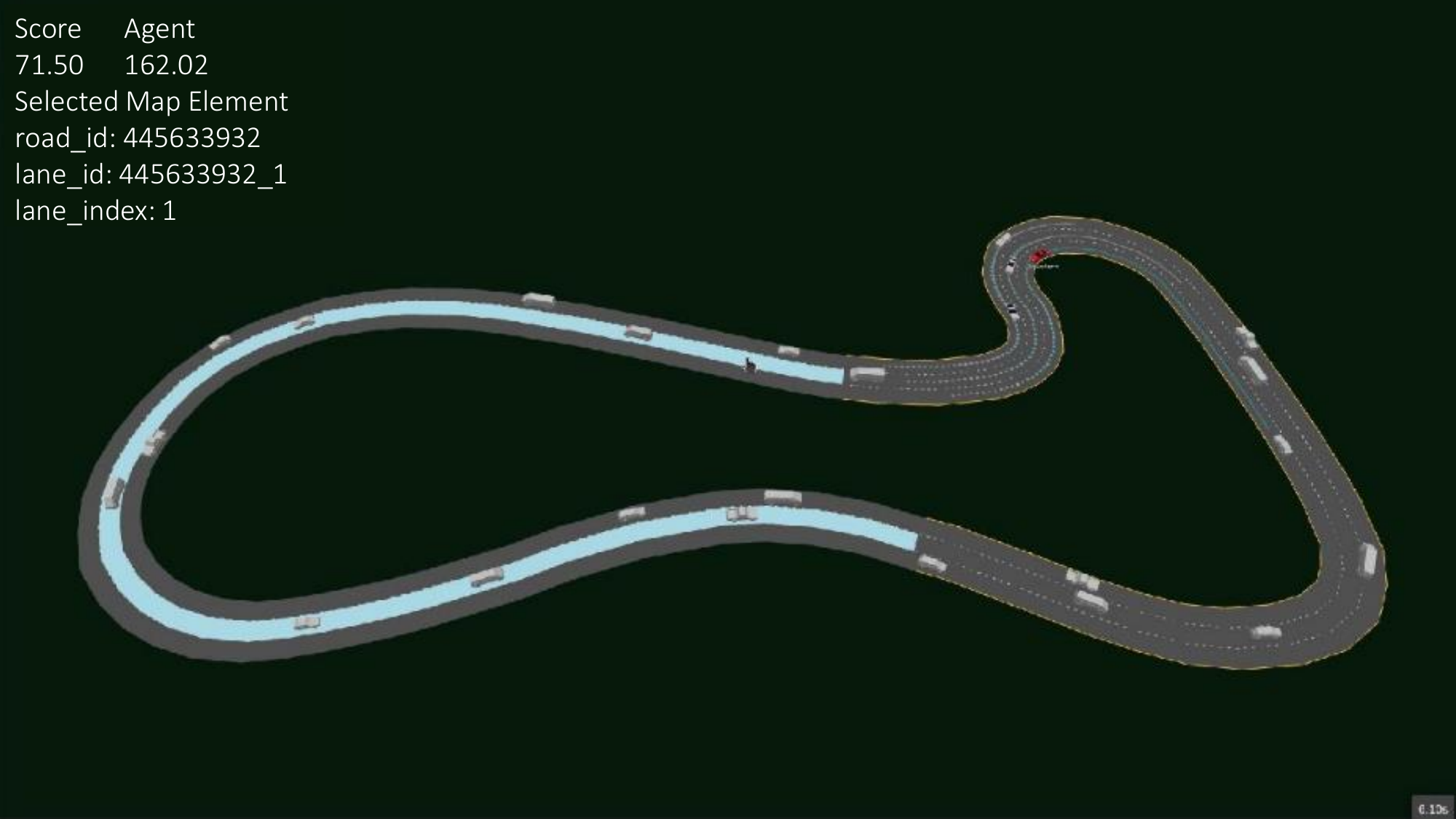}
    \caption{Overview of route selection GUI. The blue lane segment is selected by user for additional analysis. Text is enhanced from the original for better visibility.}
    \label{fig:gui}
\end{figure}

\noindent\textbf{Baseline evaluation protocol.} As part of a competition hosted in 2022 \cite{rasouli2022neurips,rasouli2023driving}, SMARTS provided an evaluation protocol for a series of common driving tasks with metrics and SOTA planning algorithms that were made online. This benchmark was fully integrated as a core feature of SMARTS 2.0.  \\
\noindent\textbf{Data integration.} Replaying naturalistic data on realistic map layouts is an effective way for simulating the behavior of social agents. Our new simulator provides support for inclusion of the map and the behavior of traffic participants from major AD datasets, such as Waymo \cite{Sun_2020_CVPR}, Argoverse \cite{Chang_2019_CVPR}, and NGSIM \cite{kovvali2007video}, and includes new support for simulating traffic light control and pedestrian agents. The agents from these datasets can be co-played with various simulated social agents. Meanwhile, there is flexibility that any replay agent can be converted to a reactive social agent.\\
\noindent\textbf{Sensor simulation.} Given the main focus of SMARTS is on accurate behavior modeling and planning, we follow a similar approach as \cite{vinitsky2022nocturne} and simulate a 2D observation of the mission vehicle's sensory view of visible areas regarding the occlusions in the scene (see Figure \ref{fig:sensor}). This approach enables the generation of local planning maps, which are becoming more common in AD research, as opposed to high definition (HD) maps, which are commonly used in other simulators.\\
\noindent\textbf{Diagnostic GUI support.} SMARTS 2.0 also provides GUI support that allows highlighting and selection of agents or lane segments. This is an effective tool for analyzing and diagnosing models' performance (see Figure \ref{fig:gui}).\\
\noindent\textbf{V2V communication.} Vehicle-to-vehicle (V2V) communication as a method for collective planning is one of the key research problems in the intelligent transportation domain \cite{xu2022opv2v}. In SMARTS 2.0, we provide a configurable base communication protocol, enabling the evaluation of cooperative planning algorithms.\\
\noindent\textbf{Real-time execution.} Although SMARTS was built with the goal of scalability for complex simulations, its performance was significantly hindered when simulating a large number of agents. In SMARTS 2.0, a series of optimizations are performed to guarantee real-time execution of planning algorithms in large cluttered environments.

\section{Interactive Motion Planning Benchmark}
\subsection{Scenarios}
We conduct the benchmarking of interactive motion planning tasks in two categories: collaborative planning and adaptive planning.
\subsubsection{Collaborative planning}
\begin{figure*}
\centering
\includegraphics[width=1\textwidth]{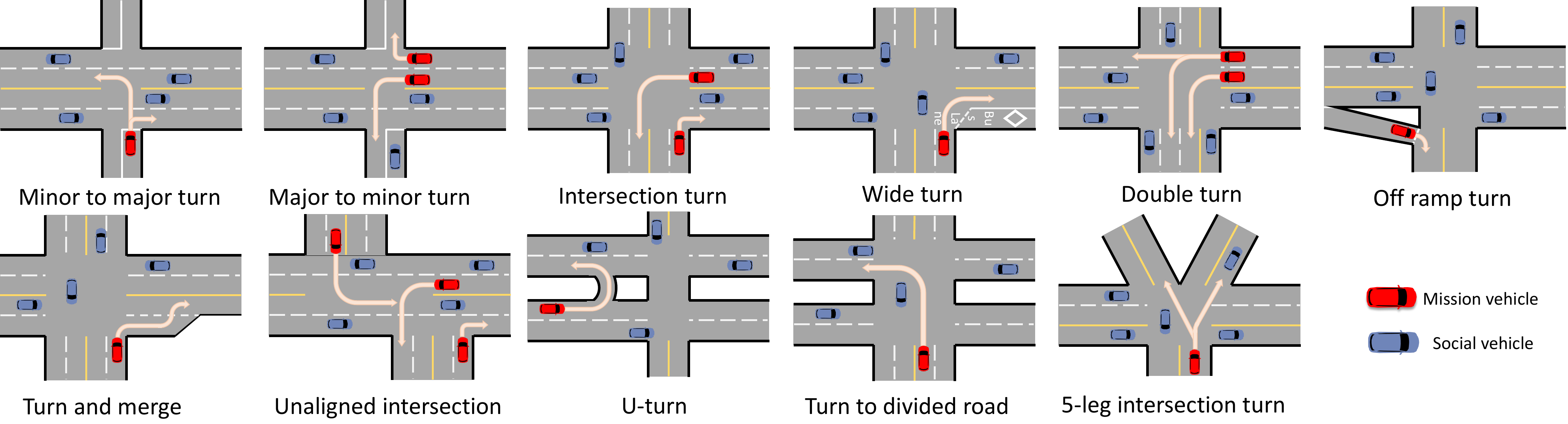}
\caption{Various types of turn maneuvers for collaborative planning.}
\label{fig:turn_scenarios}
\end{figure*}

This task category requires individual agents to negotiate scenarios with other traffic agents, to achieve their objective. In particular, we focus on the interactive scenarios at intersections for which the agents should be able to accurately model interactions between other road users and forecast their behavior. Variations within this task category's scenarios are determined by road structure, agent quantity, traffic density, presence of signals, and types of turning maneuvers. Sample turning scenarios are shown in Figure \ref{fig:turn_scenarios}.

\subsubsection{Adaptive planning}
\begin{figure*}
\centering
\includegraphics[width=1\textwidth]{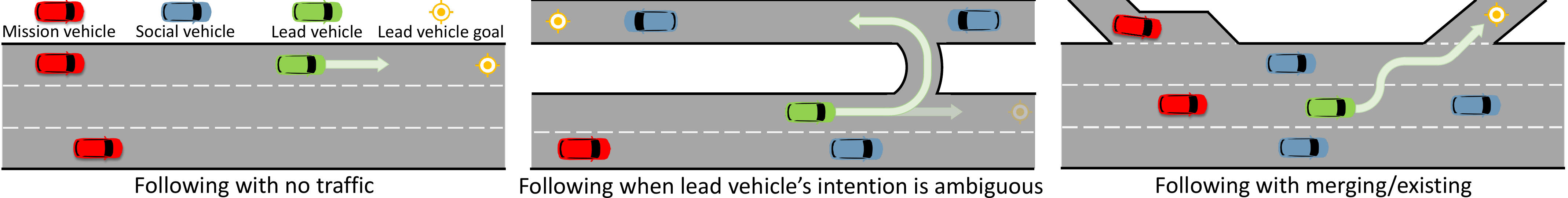}
\caption{Vehicle-following scenarios for adaptive planning.}
\label{fig:following_scenarios}
\end{figure*}

This problem addresses a key research challenge in AD where the task of the mission vehicle is to follow another vehicle and adapt to its behavior without knowing the lead vehicle's current intention and goals. The focus is on accurate sensing, accurate estimation of the intention of the lead vehicle, adaptation, and efficiency in terms of maintaining the minimum safe distance while following a lead vehicle. This task category promotes the use of multi-agent planning algorithms.

Vehicle following scenarios are likewise designed to consider varying levels of difficulty. In the simplest case, the focus of the task is on the ability of the mission vehicle to mimic the behavior of a lead vehicle with no background social vehicles present. More complex scenarios involve different numbers of social agents, the lead vehicle performing challenging behaviors (e.g., merging, exiting, turning), different starting locations for the mission vehicles, and environments that make the lead vehicle's intention ambiguous to the mission vehicles. These scenarios are depicted in Figure \ref{fig:following_scenarios}.

\subsection{Evaluation Metrics}
\label{eval_metrics}
The proposed metrics are designed to evaluate \textit{safety}, \textit{efficiency}, \textit{human-likeliness} (humanness), and \textit{completion} of the solutions. The final ranking is computed as a weighted average of the individual metrics, $S_{final} = \alpha_1 m_1 + ..., \alpha_n m_n$ where $S_{final} \in [0,1]$ and $\alpha$ and $m$ are weights and values of metrics respectively.

The metrics are divided into \textbf{common metrics} which are used for all task categories and \textbf{task-specific metrics} which are used for the corresponding scenario-specific tasks.

\subsubsection{Common Metrics}
\textbf{Progress Rate (PR)} intends to evaluate how far a mission vehicle advances towards the goal before the termination of a scenario. PR is defined as:
\begin{align*}
   & \texttt{PR} = 1 - \frac{1}{N_{sc}} \sum_{k=1}^{N_{sc}}  \frac{1}{N_{a,k}}\sum_{i=1}^{N_{a,k}}\\
   & \frac{min(dist(mv_i,g_f),dist(mv_{i-init}, g_f))}{dist(mv_{i-init}, g_f)} \in [0,1]
\end{align*}
\noindent where $N_{a,k}$ is the number ($a$) of mission vehicles ($mv$) in scenario $k$, $N_{sc}$ is the number of scenarios, $mv_{i-init}$ is the initial position of the mission vehicle $i$, $mv_i$ is the final position of the mission vehicle at termination, and $g_f$ is the position of the final goal at termination time.

\begin{figure}
\centering
\includegraphics[width=0.7\columnwidth]{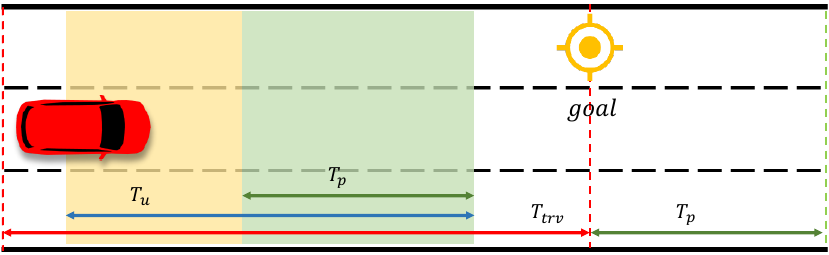}
\caption{Driving comfort $comf$ based on travel time to the destination ($T_{trv}$), uncomfortable period ($T_u$), and penalty period ($T_p$).}
\label{fig:human_metric}
 \vspace{-0.3cm}
\end{figure}

\textbf{Humanness} is intended to approximate how human-like the behavior of the mission vehicles is, motivated by an existing proposal \cite{bae2020self}. This metric is based on the level of comfort (due to changes in acceleration and jerk) and lane center offset formulated as follows (see Figure \ref{fig:human_metric}):

\begin{align*}
&\texttt{Humanness} =  1 - \frac{1}{N_{sc}} \sum_{k=1}^{N_{sc}}\\
& \frac{1}{2N_{a,k}} 
\sum_{i=1}^{N_{a,k}}
\left(comf_{k,i} + 
\frac{1}{T_{sc_{i,k}}} 
\sum_{T_{sc_{i,k}}}
\texttt{lc-off}_{i}\right) \in [0,1]
\end{align*}
\noindent where $T_{sc_{k,i}} = min(T_{mv_{i,k}},T_{sc_k})$. Here, $T_{mv_{i,k}}$ is the time for vehicle $i$ to complete scenario $k$ and $T_{sc_k}$ is time limit of scenario $k$, $\texttt{lc-off}$ is lane center offset, which is $1$ if the vehicle is fully $\texttt{offroad}$ when the majority of a vehicle's wheels have left the road surface, otherwise given by $\frac{dist(mv,lc)}{lw*0.5}$, where $lw$ denotes the lane width. $comf$ metric is computed by,

\begin{align*}
& comf = \frac{T_u}{T_{trv} + T_p}, \texttt{ }\\
& T_u = \sum_{i=1}^{T_{trv}+T_p} u_{t_i},
u_t=\left\{\begin{matrix}
1 & \underset{\alpha\leqslant t'\leqslant \beta }{max} dyn_{t'} > 1\\ 
0 & \texttt{otherwise}
\end{matrix}\right.
\end{align*}

\noindent where $\alpha = max(0, t-T_p)$, $\beta = min(T_{trv},t)$, and $dyn =  max\left(\frac{(\texttt{jerk}_{\texttt{lin}}, \texttt{acc}_{lin})}{(\texttt{jerk}_{\texttt{lin-max}}, \texttt{acc}_{lin-max})}\right)$ where $\texttt{jerk}_{lin-max} = (0.9,0.9)$ and $\texttt{acc}_{lin-max} = (2.0,1.47)$ referring to $(\texttt{long}, \texttt{lat})$ values for maximum linear jerk and acceleration respectively \cite{bae2020self}. 

\textbf{Rule Compliance (RC)} measures how safe the behavior of the vehicle is in terms of following road rules. The RC metric is specified as follows:

\begin{align*}
& \texttt{RC} = 1 - \frac{1}{N_{sc}} 
\sum_{k=1}^{N_{sc}} 
\frac{1}{3N_{a,k}}\\
&\left(
\sum_{i=1}^{N_{a,k}} 
\frac{1}{T_{sc_{k,i}}}
\sum_{T_{sc_{k,i}}} 
min(\frac{s_{violate}}{0.5 s_{limit}},1)+r_{violate}+po \right)\in [0,1]
\end{align*}
 
\noindent where $r_{violate} \in \{0,1\}$ indicates whether the vehicle violated road direction, $s_{limit}$ is speed limit, and $s_{violate}$ is the amount of violated speed and $po \in \{0,1\}$ is a partial \texttt{offroad} that indicates whether at least one wheel of the mission vehicle is out of the road boundary.

\subsubsection{Task-specific Metrics}
 \textbf{Mission Time Efficiency (MTE)} is based on when the vehicles arrive at their destinations and evaluate the time efficiency of navigation. The value is given by:

$$\texttt{MTE} = 1 - \frac{1}{N_{sc}} \sum_{k=1}^{N_{sc}} \frac{1}{T_{sc_k}
N_{a,k}}\sum_{i=1}^{N_{a,k}} Tr_{mv_i} \in [0,1],$$
\noindent where $Tr_{mv_i}$ is the time for vehicle $i$ to arrive within the goal arrival radius.

 \textbf{Safe Following Distance (SFD)} is a metric specific to adaptive planning that evaluates how well the mission vehicle associates itself to the safe follow margin behind the lead vehicle during the vehicle following task (see Figure \ref{fig:SFD_metric}). SFD is defined as:
 \vspace{-0.3cm}

\begin{figure}
\centering
\includegraphics[width=0.7\columnwidth]{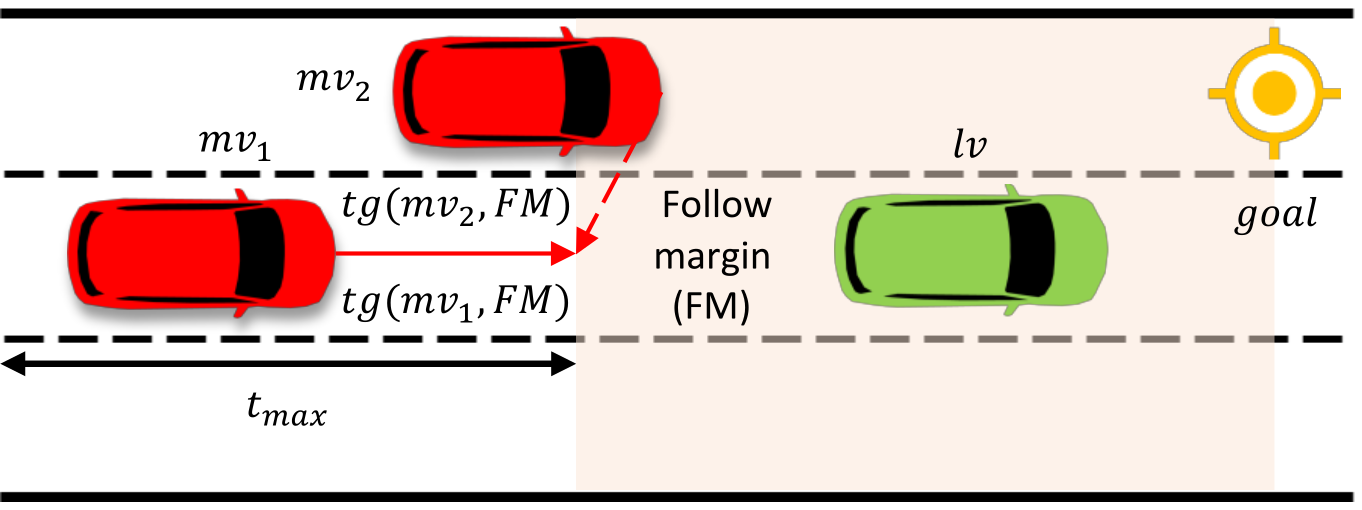}
\caption{Safe Following Distance (SFD) in seconds based on absolute distance to safe follow margin around the lead vehicle.}
\label{fig:SFD_metric}
 \vspace{-0.3cm}
\end{figure}

 $$ \texttt{SFD} = 1 - \frac{1}{N_{sc}}
\sum_{k=1}^{N_{sc}}
 \frac{1}{N_{a,k}}
 \sum_{i=1}^{N_{a,k}}
 \frac{1}{T_{sc_{i,k}}}
 \sum_{T_{sc_{i,k}}}
 mv_{i-gap} \in [0,1],$$
 
 $$mv_{i-gap}=\left\{\begin{matrix}
1 & \texttt{if } mv_i  \in FM \texttt{ or } \\
&tg(mv_i, FM) > t_{max} \\ 
 \frac{ tg(mv_i, FM)}{t_{max}}& \texttt{otherwise}
\end{matrix}\right..$$
Here, $t_{max}$ is the maximum acceptable time range to follow margin ($FM$) and $tg(mv_i, FM)$ is the time gap of vehicle $i$ to $FM$ in seconds.

\section{Evaluation}
\subsection{SMARTS 2.0 Performance}
 In this section, we conduct a performance test between SMARTS 2.0 and its predecessor, SMARTS\;\cite{zhou2020smarts} using three evaluation criteria: the number of Integrated Traffic Actors, the number of Agents, and Road Network Edges. \textbf{Integrated Traffic Actors} evaluates the performance of the simulation given a quantity of internal traffic simulator vehicles to simulate. \textbf{Agents} measures the simulation of the number of agent vehicles, given road and neighboring vehicle sensors. \textbf{Road Network Edges} measures the simulation performance interaction between 10 agents and a scaling quantity of road network edges in the road network.
 
 The performance is measured using the inbuilt SMARTS diagnostic tool running on a device with 32GB RAM and an Intel(R) Core(TM) i9-9900K CPU with 3.60GHz clock speed. As shown in Table \ref{tab:performance-12}, SMARTS 2.0 clearly stands out in most conditions, gaining as high as 835\% improvement compared to SMARTS. In particular, SMARTS 2.0 is more effective for multi-agent simulation and large environments, offering 200\%+ across all settings.

\begin{table}[]
\caption{From left to right: the name of the test variables(\textbf{Test Var.}), the value (number) of the evaluated test variables (\textbf{No.}), the frames per second (FPS) of SMARTS 1.0 (\textbf{S1.0}), the FPS of SMARTS 2.0 on the same diagnostic group (\textbf{S2.0}), and the relative difference in performance upgrading to SMARTS 2.0 (\textbf{Diff.}). Each FPS value is averaged over 10,000 steps.}
\centering
 \resizebox{\columnwidth}{!}{\begin{tabular}{c |c c c |c }
\hline
\multicolumn{1}{l|}{\textit{Test Var.}}                      & \textit{No.}                  & \textit{S1.0}& \textit{S2.0 }& \textit{Diff.}   \\ \hline
    & 1                     & 1544                 & 1301                 & -16\% \\
 & 10                    & 1366                 & 988                  & -28\% \\
  & 20                    & 1317                 & 1055                 & -20\% \\
     
\multirow{-5}{1.7cm}{\textbf{Integrated Traffic Actors}} & 50                    & 37                   & 346                  & 835\% \\\hline
                                    & 1                     & 401                  & 1243                 & 210\% \\
                                    & 10                    & 230                  & 1188                 & 418\% \\
                                    & 20                    & 233                  & 1178                 & 405\% \\
\multirow{-4}{1.7cm}{\textbf{Agent}}                       & 50                    & 134                  & 1157                 & 762\% \\ \hline
                                         & 1     & 205                  & 1369                & 567\% \\
                                          & 10     & 202                  & 1361                 & 573\% \\
                                             & 20     & 202                  & 1379                 & 581\% \\
\multirow{-4}{1.7cm}{\textbf{Road Network Edges}} & 50     & 198                  & 1399                 & 604\% \\ \hline
\end{tabular}}
\label{tab:performance-12}
\vspace{-0.4cm}
\end{table}

\subsection{The Benchmark}
\textbf{Metrics.} We use the metrics as described in Section \ref{eval_metrics}. In addition, for ranking of complete alternative approaches we resort to a combined metric $S_{bench}$,
\[
S_{\text{bench}} = \frac{\sum_{i=1}^{n} w_i \, M_i}{\sum_{i=1}^{n} w_i} \in [0, 1], n = {\sum_{i=1}^{n} w_i},
\]
\noindent where $M$ is an evaluation metric result and $w$ is the scoring weight of evaluation metric $i$. We empirically set the weights as $0.1$ for PR, $0.45$ for RC, $0.15$ for Humanness, and $0.3$ each for MTE and SFD in each benchmark task.

\textbf{Methods.} For evaluation, we use the three winning methods from our 2022 competition, namely \textbf{TF}\footnote{\href{https://github.com/superCat-star/fanta-code}{https://github.com/superCat-star/fanta-code}}, \textbf{VCR}\footnote{\href{https://github.com/yuant95/SMARTS_VCR},{https://github.com/yuant95/SMARTS\_VCR}}, and \textbf{AID}\footnote{\href{https://github.com/MCZhi/Predictive-Decision}{https://github.com/MCZhi/Predictive-Decision}}. TF is a hierarchical model with a meta controller and a scheduler. The controller consists of a rule-based collision detector that recognizes situations in which an incident might take place, and an offline learning agent that generates actions. The scheduler executes the decision policy provided by the controller using a series of underlying policies, including speed, moving direction, and merging. 

VCR method uses two controllers, namely baseline and filtering. The former relies on the waypoints provided in the observation space, from which the controller selects the ones that lead to the goal location. The filtering controller, then samples from the lines between the agents current location and next, and using a neural network, ranks them based on their probability of collisions or other termination events. Lastly, AID model employs a hybrid transformer-based approach consisting of a motion predictor and sampling-based planner. The predictor forecasts the future trajectories of surrounding social agents and the planner selects the optimal trajectory considering the distance to the mission goal, comfort, and safety (more details can be found in \cite{rasouli2022neurips}). The three SOTA models are benchmarked using pre-existing submission checkpoints without modifications.

We additionally evaluate three baseline models (which are available in SMARTS), \textbf{Drive} for collaborative planning, and \textbf{Platoon-c} and \textbf{Platoon-rtp} for adaptive planning. These reinforcement learning (RL) based approaches use the PPO \cite{schulman2017proximal} algorithm. Their policies share the same components but 
they differ in how they interpret the observations from the environment to get a reward signal for their associated tasks.

\textbf{Implementation.} All models receive the same sensor observations about the environment. The benchmark provides three different action spaces for policies and does not inherently discriminate result categories for each. These action spaces affect the action that the agent may choose, and the way that the agent's mission vehicle is constrained by its repertoire of possible actions within the task environment.

\noindent\textbf{Relative Target Pose}. The agent can select its next position and heading relative to its current state, without direct regard to physics simulation. The benchmark then scales back the action to a range that is possible for the vehicle. The models Drive and Platoon-rtp use this action space.

\noindent\textbf{Target Pose}. The agent can select its mission vehicle's next position and heading relative to its state in the global frame. The benchmark then scales back the action to a range that is possible for the vehicle in a way that makes the action equivalent to the Relative Target Pose action space. The competition previously conducted captures  TF, VCR, and AID use this action space.

\noindent\textbf{Continuous}. The agent uses throttle, steering, and break to control a physics engine simulated mission vehicle. The state transition of the vehicle is limited by physics forces within the simulation. Platoon-c uses this action space.

\begin{figure*}
    \centering
    \includegraphics[width=1\textwidth]{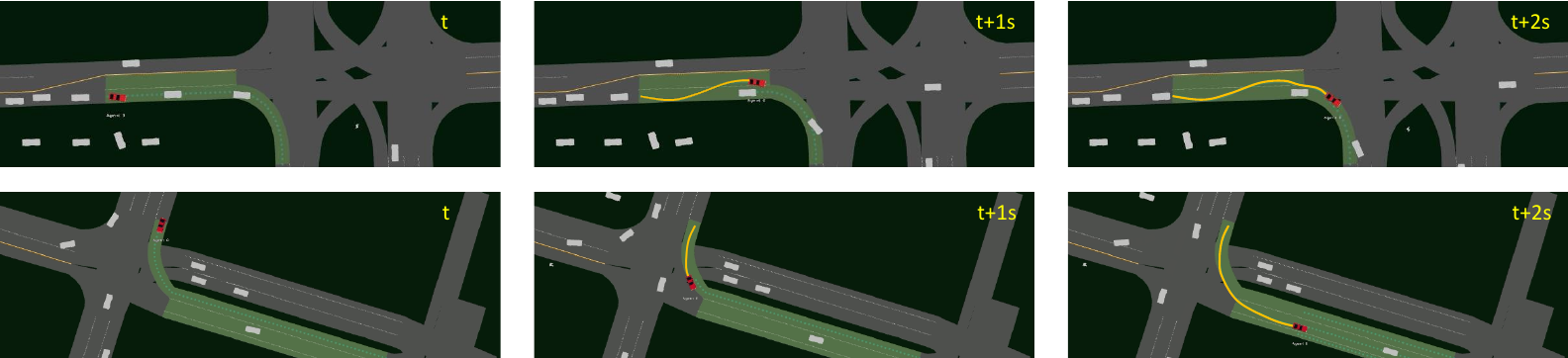}
    \caption{Examples of right turn with cut in (\textbf{top}) and left turn with lane change (\textbf{bottom}) using Drive model. The {\color{red}mission vehicle} and  {\color{gray}social agents} are shown in a scenario played in an environment from Argoverse. {\color{Dandelion}Driving trajectory} is enhanced for better visibility.}
    \label{fig:turn_exp}
    \vspace{-0.3cm}
\end{figure*}

\begin{figure*}
    \centering
    \includegraphics[width=1\textwidth]{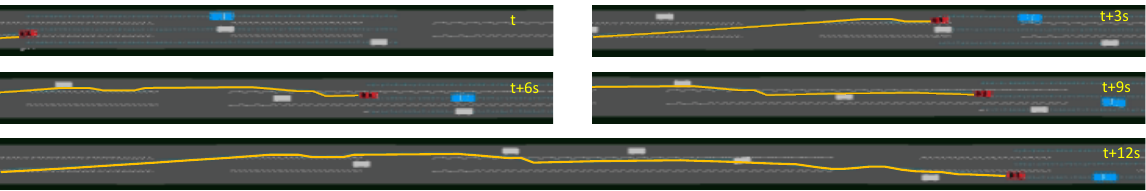}
    \caption{Example of car following scenario at different timesteps using Platoon-c model. The {\color{red}mission vehicle}, {\color{cyan} lead vehicle}, and  {\color{gray}social agents} are shown in a scenario played in a synthesized environment. {\color{Dandelion}Driving trajectory} is enhanced for better visibility.}
    \label{fig:follow_exp}
    \vspace{-0.4cm}
\end{figure*}

\subsubsection{Collaborative planning}
\begin{table}[]
\caption{Benchmark results for collaborative planning tasks. For all values, higher is better.}
 \resizebox{\columnwidth}{!}{\begin{tabular}{l|cccc|c}
\hline
\textit{Method} & \textit{PR}    & \textit{RC}    & \textit{Humanness} & \textit{MTE}   & \textit{S$_{bench}$} \\ \hline
\textbf{AID}    & 0.895          & 0.776          & 0.618             & \textbf{0.221}          & \textbf{0.598}   \\  
\textbf{VCR}    &\textbf{0.958} & \textbf{0.789}        & 0.532             & 0.195          & 0.589            \\
\textbf{Drive}  & 0.715         & 0.669          & \textbf{0.917}    & 0.178& 0.564  \\
\textbf{TF}     & 0.956         & 0.744 & 0.620             & 0.130          & 0.562            \\
\end{tabular}}
\label{tab:benchmark-cp}
\vspace{-0.3cm}
\end{table}
The results of the collaborative benchmark are reported in Table \ref{tab:benchmark-cp}. At first glance, we can see that the performances of the models vary on different metrics. While VCR achieves the best performance on PR and RC, Drive performs best on Humanness, and AID on MTE. All models perform poorly on the MTE metric, pointing to the fact that they are overly conservative.

In terms of discrepancy, the biggest gap can be observed on Humanness metric. Here, Drive is significantly better as its reward encourages the agent to stay within its lane, whereas other models emphasize task completion, and as a result, do better on the PR metric. Overall, AID achieves the best performance on $S_{bench}$ metric as it achieves the close second on the core RC metric and the best on MTE as the primary task-based metric. 

\subsubsection{Adaptive Planning}

\begin{table}[]
\caption{Benchmark results for adaptive planning tasks. For all values, higher is better.}
 \resizebox{\columnwidth}{!}{\begin{tabular}{l|cccc|c}
\hline
\textit{Method}      & \textit{PR}    & \textit{RC}    & \textit{Humanness} & \textit{SFD}   & \textit{S$_{bench}$} \\ \hline
\textbf{Platoon-rtp} & 0.438       & 0.855  & \textbf{0.902}    & \textbf{0.361}         & \textbf{0.672}           \\
\textbf{Platoon-c}   &   0.671   & \textbf{0.917}    & 0.444     & 0.182          & 0.601  \\

\textbf{VCR}   &   \textbf{0.822}   & 0.708   & 0.546  &  0.050    &  0.498           \\
\textbf{AID}         & 0.802 & 0.695& 0.759              & 0.070          & 0.528  \\
\textbf{TF}          & 0.773         & 0.711  & 0.727             & 0.010          & 0.509
\end{tabular}}
\label{tab:benchmark-ap}
\vspace{-0.4cm}
\end{table}

As shown in Table \ref{tab:benchmark-ap}, once again we observe diverse performance variations across different metrics. VCR performs best on PR but the baseline platoon models perform poorly on this metric, which indicates that they deviate significantly from the path of the lead vehicle. However, Platoon-c achieves the best performance on RC but at the cost of performing worst in Humanness. Platoon-rtp, on the other hand, achieves the best performance on Humanness and SFD, making this model the best on the combined metric, $S_{bench}$.

\subsubsection{Qualitative Samples}
\textbf{Collaborative driving} samples using Drive are shown in Figure \ref{fig:turn_exp}. In these examples, we see the complexity of the proposed scenarios and resulted erratic behaviors of the model. In the right turn scenario (top), the vehicle performs a cut-in action while attempting to make a right turn. This can be due to the fact that the model is prioritizing reaching goal over safety and comfort factors. In the left turn example (bottom), the vehicle drifts to far right lane before completing its turn, pointing to the fact that the model is maximizing humanness factor over regulation. As shown in the examples, balancing the performance with respect to different metrics can be quite challenging.

\textbf{Vehicle following } results in a scenario using Platoon-c method is illustrated in Figure \ref{fig:follow_exp}. Here, we can see that even though the mission vehicle successfully follows the lead vehicle, its path is quite inconsistent, unhuman-like, and somewhat erratic. This once again illustrates the importance of benchmarking comprehensive aspects of motion planning, in this case, humanness.

\section{Conclusion}
In this paper, we introduced SMARTS 2.0, an extension of our widely used simulation platform that offers many enhanced features, such as real data replay, advanced sensor simulation, V2V communication, and diagnostic tools in a highly optimized framework. In addition, we proposed a novel benchmark for highly collaborative planning tasks involving challenging intersections, and adaptive planning consisting of the vehicle following task. We conducted experiments using baseline and SOTA algorithms using our proposed metrics, and show that all approaches struggle to output a balanced performance across all metrics, making their applicability to real-world situations questionable. For future work, one can consider revisiting metrics such as RC to impose more restrictive conditions better resembling real-world constraints. As for future scenario improvements, inclusion of connected-driving settings requiring V2V communication, variations in traffic density, and heterogeneity can be further considered.
\bibliographystyle{IEEEtran}
\bibliography{references}

\begin{thebibliography}{10}
\providecommand{\url}[1]{#1}
\csname url@rmstyle\endcsname
\providecommand{\newblock}{\relax}
\providecommand{\bibinfo}[2]{#2}
\providecommand\BIBentrySTDinterwordspacing{\spaceskip=0pt\relax}
\providecommand\BIBentryALTinterwordstretchfactor{4}
\providecommand\BIBentryALTinterwordspacing{\spaceskip=\fontdimen2\font plus
\BIBentryALTinterwordstretchfactor\fontdimen3\font minus
  \fontdimen4\font\relax}
\providecommand\BIBforeignlanguage[2]{{%
\expandafter\ifx\csname l@#1\endcsname\relax
\typeout{** WARNING: IEEEtran.bst: No hyphenation pattern has been}%
\typeout{** loaded for the language `#1'. Using the pattern for}%
\typeout{** the default language instead.}%
\else
\language=\csname l@#1\endcsname
\fi
#2}}

\bibitem{zhang2022rethinking}
C.~Zhang, R.~Guo, W.~Zeng, Y.~Xiong, B.~Dai, R.~Hu, M.~Ren, and R.~Urtasun,
  ``Rethinking closed-loop training for autonomous driving,'' in \emph{ECCV},
  2022.

\bibitem{zhou2020smarts}
M.~Zhou, J.~Luo, J.~Villella, Y.~Yang, D.~Rusu, J.~Miao, W.~Zhang, M.~Alban,
  I.~Fadakar, Z.~Chen, \emph{et~al.}, ``{SMARTS: Scalable Multi-Agent
  Reinforcement Learning Training School for Autonomous Driving},'' in
  \emph{CoRL}, 2020.

\bibitem{wymann2000torcs}
\BIBentryALTinterwordspacing
B.~Wymann, E.~Espi{\'e}, C.~Guionneau, C.~Dimitrakakis, R.~Coulom, and
  A.~Sumner, ``Torcs, the open racing car simulator,'' Online, 2000. [Online].
  Available: \url{https://sourceforge.net/projects/torcs/}
\BIBentrySTDinterwordspacing

\bibitem{leurent2018environment}
\BIBentryALTinterwordspacing
E.~Leurent, ``An environment for autonomous driving decision-making,'' Online,
  2018. [Online]. Available: \url{https://github.com/eleurent/highway-env}
\BIBentrySTDinterwordspacing

\bibitem{gulino2024waymax}
C.~Gulino, J.~Fu, W.~Luo, G.~Tucker, E.~Bronstein, Y.~Lu, J.~Harb, X.~Pan,
  Y.~Wang, X.~Chen, \emph{et~al.}, ``Waymax: An accelerated, data-driven
  simulator for large-scale autonomous driving research,'' in \emph{NeurIPS},
  2024.

\bibitem{muller2018sim4cv}
M.~M{\"u}ller, V.~Casser, J.~Lahoud, N.~Smith, and B.~Ghanem, ``Sim4cv: A
  photo-realistic simulator for computer vision applications,''
  \emph{International Journal of Computer Vision}, vol. 126, pp. 902--919,
  2018.

\bibitem{dosovitskiy2017carla}
A.~Dosovitskiy, G.~Ros, F.~Codevilla, A.~Lopez, and V.~Koltun, ``{CARLA: An
  Open Urban Driving Simulator},'' in \emph{CoRL}, 2017.

\bibitem{sun2022intersim}
Q.~Sun, X.~Huang, B.~C. Williams, and H.~Zhao, ``Intersim: Interactive traffic
  simulation via explicit relation modeling,'' in \emph{IROS}, 2022.

\bibitem{scibior2021imagining}
A.~{\'S}cibior, V.~Lioutas, D.~Reda, P.~Bateni, and F.~Wood, ``Imagining the
  road ahead: Multi-agent trajectory prediction via differentiable
  simulation,'' in \emph{ITSC}, 2021.

\bibitem{martinez2017beyond}
M.~Martinez, C.~Sitawarin, K.~Finch, L.~Meincke, A.~Yablonski, and
  A.~Kornhauser, ``Beyond grand theft auto v for training, testing and
  enhancing deep learning in self driving cars,'' \emph{arXiv:1712.01397},
  2017.

\bibitem{amini2022vista}
A.~Amini, T.-H. Wang, I.~Gilitschenski, W.~Schwarting, Z.~Liu, S.~Han,
  S.~Karaman, and D.~Rus, ``Vista 2.0: An open, data-driven simulator for
  multimodal sensing and policy learning for autonomous vehicles,'' in
  \emph{ICRA}, 2022.

\bibitem{cai2020summit}
P.~Cai, Y.~Lee, Y.~Luo, and D.~Hsu, ``Summit: A simulator for urban driving in
  massive mixed traffic,'' in \emph{ICRA}, 2020.

\bibitem{palanisamy2020multi}
P.~Palanisamy, ``Multi-agent connected autonomous driving using deep
  reinforcement learning,'' in \emph{International Joint Conference on Neural
  Networks}, 2020, pp. 1--7.

\bibitem{craig_quiter_2020}
C.~Quiter, ``Deepdrive zero,'' June 2020.

\bibitem{santara2021madras}
A.~Santara, S.~Rudra, S.~A. Buridi, M.~Kaushik, A.~Naik, B.~Kaul, and
  B.~Ravindran, ``Madras: Multi agent driving simulator,'' \emph{Journal of
  Artificial Intelligence Research}, vol.~70, pp. 1517--1555, 2021.

\bibitem{althoff2017commonroad}
M.~Althoff, M.~Koschi, and S.~Manzinger, ``{CommonRoad: Composable Benchmarks
  for Motion Planning on Roads},'' in \emph{IV}, 2017.

\bibitem{caesar2021nuplan}
H.~Caesar, J.~Kabzan, K.~S. Tan, W.~K. Fong, E.~Wolff, A.~Lang, L.~Fletcher,
  O.~Beijbom, and S.~Omari, ``nuplan: A closed-loop ml-based planning benchmark
  for autonomous vehicles,'' \emph{arXiv:2106.11810}, 2021.

\bibitem{li2022metadrive}
Q.~Li, Z.~Peng, L.~Feng, Q.~Zhang, Z.~Xue, and B.~Zhou, ``Metadrive: Composing
  diverse driving scenarios for generalizable reinforcement learning,''
  \emph{TPAMI}, vol.~45, no.~3, pp. 3461--3475, 2022.

\bibitem{vinitsky2022nocturne}
E.~Vinitsky, N.~Lichtl{\'e}, X.~Yang, B.~Amos, and J.~Foerster, ``Nocturne: a
  scalable driving benchmark for bringing multi-agent learning one step closer
  to the real world,'' in \emph{NeurIPS}, 2022.

\bibitem{xu2023bits}
D.~Xu, Y.~Chen, B.~Ivanovic, and M.~Pavone, ``Bits: Bi-level imitation for
  traffic simulation,'' in \emph{ICRA}, 2023.

\bibitem{Li_2024_tiv}
Y.~Li, W.~Yuan, S.~Zhang, W.~Yan, Q.~Shen, C.~Wang, and M.~Yang, ``Choose your
  simulator wisely: A review on open-source simulators for autonomous
  driving,'' \emph{Transactions on Intelligent Vehicles}, vol.~9, no.~5, pp.
  4861--4876, 2024.

\bibitem{houston2021one}
J.~Houston, G.~Zuidhof, L.~Bergamini, Y.~Ye, L.~Chen, A.~Jain, S.~Omari,
  V.~Iglovikov, and P.~Ondruska, ``One thousand and one hours: Self-driving
  motion prediction dataset,'' in \emph{CoRL}, 2021.

\bibitem{caesar2020nuscenes}
H.~Caesar, V.~Bankiti, A.~H. Lang, S.~Vora, V.~E. Liong, Q.~Xu, A.~Krishnan,
  Y.~Pan, G.~Baldan, and O.~Beijbom, ``{nuScenes: A Multimodal Dataset for
  Autonomous Driving},'' in \emph{CVPR}, 2020.

\bibitem{Wilson_Argoverse2}
B.~Wilson, W.~Qi, T.~Agarwal, J.~Lambert, J.~Singh, S.~Khandelwal, B.~Pan,
  R.~Kumar, A.~Hartnett, J.~K. Pontes, D.~Ramanan, P.~Carr, and J.~Hays,
  ``Argoverse 2: Next generation datasets for self-driving perception and
  forecasting,'' in \emph{NeurIPS}, 2021.

\bibitem{chen2024criteria}
C.~Chen, M.~Pourkeshavarz, and A.~Rasouli, ``Criteria: a new benchmarking
  paradigm for evaluating trajectory prediction models for autonomous
  driving,'' in \emph{ICRA}, 2024.

\bibitem{rasouli2023driving}
A.~Rasouli, S.~Alizadeh, I.~Kotseruba, Y.~Ma, H.~Liang, Y.~Tian, Z.~Huang,
  H.~Liu, J.~Wu, R.~Goebel, \emph{et~al.}, ``Driving smarts competition at
  neurips 2022: Insights and outcome,'' in \emph{NeurIPS}, 2023.

\bibitem{Sun_2020_CVPR}
P.~Sun, H.~Kretzschmar, X.~Dotiwalla, A.~Chouard, V.~Patnaik, P.~Tsui, J.~Guo,
  Y.~Zhou, Y.~Chai, B.~Caine, \emph{et~al.}, ``{Scalability in Perception for
  Autonomous Driving: Waymo Open Dataset},'' in \emph{CVPR}, 2020.

\bibitem{Ansys_comp}
\BIBentryALTinterwordspacing
``{Ansys Indy Autonomous Challenge Simulation Race},'' Online, 2021. [Online].
  Available: \url{https://www.ansys.com/it-it/experiences/iacrace}
\BIBentrySTDinterwordspacing

\bibitem{TPCAP_comp}
\BIBentryALTinterwordspacing
B.~Li, X.~Wang, S.~Li, L.~Li, and F.-Y. Wang, ``{TPCAP: Trajectory Planning
  Competition for Automated Parking},'' Online, 2022. [Online]. Available:
  \url{https://www.tpcap.net/\#/}
\BIBentrySTDinterwordspacing

\bibitem{rasouli2022neurips}
A.~Rasouli, R.~Goebel, M.~E. Taylor, I.~Kotseruba, S.~Alizadeh, T.~Yang,
  M.~Alban, F.~Shkurti, Y.~Zhuang, A.~Scibior, \emph{et~al.}, ``{NeurIPS 2022
  Competition: Driving SMARTS},'' \emph{arXiv:2211.07545}, 2022.

\bibitem{Chang_2019_CVPR}
M.-F. Chang, J.~Lambert, P.~Sangkloy, J.~Singh, S.~Bak, A.~Hartnett, D.~Wang,
  P.~Carr, S.~Lucey, D.~Ramanan, and J.~Hays, ``Argoverse: {3D} tracking and
  forecasting with rich maps,'' in \emph{CVPR}, 2019.

\bibitem{kovvali2007video}
V.~G. Kovvali, V.~Alexiadis, and L.~Zhang, ``Video-based vehicle trajectory
  data collection,'' in \emph{Transportation Research Board Annual Meeting},
  2007.

\bibitem{xu2022opv2v}
R.~Xu, H.~Xiang, X.~Xia, X.~Han, J.~Li, and J.~Ma, ``Opv2v: An open benchmark
  dataset and fusion pipeline for perception with vehicle-to-vehicle
  communication,'' in \emph{ICRA}, 2022.

\bibitem{bae2020self}
I.~Bae, J.~Moon, J.~Jhung, H.~Suk, T.~Kim, H.~Park, J.~Cha, J.~Kim, D.~Kim, and
  S.~Kim, ``Self-driving like a human driver instead of a robocar: Personalized
  comfortable driving experience for autonomous vehicles,''
  \emph{arXiv:2001.03908}, 2020.

\bibitem{schulman2017proximal}
J.~Schulman, F.~Wolski, P.~Dhariwal, A.~Radford, and O.~Klimov, ``Proximal
  policy optimization algorithms,'' \emph{arXiv:1707.06347}, 2017.

\end{thebibliography}

\end{document}